\documentclass[12pt]{article}

\usepackage{setspace} 
\usepackage{amsmath}
\usepackage{graphicx}
\usepackage{natbib}
\usepackage{booktabs,siunitx}
\usepackage{multirow}
\usepackage{amsfonts}
\usepackage{hyperref}
\usepackage{breakurl}

\allowdisplaybreaks[4]
\newcommand{\R}{\mathbb{R}}

\newtheorem{theorem}{Theorem}[section] 
\newtheorem{definition}[theorem]{Definition}
\newtheorem{lemma}[theorem]{Lemma}
\newtheorem{property}[theorem]{Property}

\title{Validation-Free Sparse Learning: A Phase Transition Approach to Feature Selection}

\author{Maxime van Cutsem\\ Department of Mathematics, University of Geneva\\ \texttt{Maxime.Vancutsem@unige.ch}\\
Sylvain Sardy\\ Department of Mathematics, University of Geneva\\ \texttt{sylvain.sardy@unige.ch}\\
Xiaoyu Ma \\ National University of Defense Technology\\
\texttt{xyu.ma@outlook.com} 
}

\begin{document}

\maketitle

\begin{abstract}
    The growing environmental footprint of artificial intelligence (AI), especially in terms of storage and computation, calls for more frugal and interpretable models. Sparse models (e.g., linear, neural networks) offer a promising solution by selecting only the most relevant features, reducing complexity, preventing over-fitting and enabling interpretation-marking a step towards truly intelligent AI.
    
    The concept of a {\it right} amount of sparsity (without too many false positive or too few true positive) is subjective. 
    So we propose a new paradigm previously only observed and mathematically studied for compressed sensing (noiseless linear models): obtaining a phase transition in the probability of retrieving the relevant features. We show  in practice how to obtain this phase transition for a class of sparse learners. 
    Our approach is flexible and applicable to complex models ranging from linear to shallow and deep artificial neural networks while supporting various loss functions and sparsity-promoting penalties. It does not rely on cross-validation or on a validation set to select its single regularization parameter.
    For  real-world data, it provides a good balance between predictive accuracy and feature sparsity.

    A Python package is available at https://github.com/VcMaxouuu/HarderLASSO containing all the simulations and ready-to-use models.

\end{abstract}


\section{Introduction}

\subsection{Over‐parameterization and the case for sparsity}

In machine learning, models ranging from linear models to artificial neural networks (ANNs) have achieved a remarkable success, the former for its simplicity, the latter for its flexibility (dense in smooth functions spaces \citep{Chen:1995:UAN:2325866.2328543}) and its ability to train well from data.
However, high dimensional linear models and ANNs are  over-parametrized, with multiple distinct settings of the parameters leading to the same prediction. Their large number of parameters and deep layered structure make them essentially uninterpretable.  Over-parametrization calls for regularization, for example, dropout \citep{DBLP:journals/jmlr/SrivastavaHKSS14}, early stopping, stochastic gradient, ridge \citep{ridgeHK}, to name a few.

In light of Occam's razor, \textit{"Entities must not be multiplied beyond necessity"}, among all the models with similar predictive capability, the one with the smallest number of parameters should be selected. Yet in practice, the focus has shifted almost exclusively to predictive performance, at the expense of energy efficiency and financial cost due to the excessive use of parameters. 

Sparse learners aim to simplify complex models by selecting only a small subset of inputs (features). This regularization approach is motivated by two main goals. The first is \textit{better generalization}: by focusing only on the most informative features, the model is less prone to overfitting and is more likely to perform well on unseen data. The second is \textit{interpretability}: selecting only a small subset of features makes the model's behavior and the dataset more understandable, allowing a human scientist to give an intelligent interpretation to the few selected features. For example, if genes are the inputs and cancer type is the output, identifying a small set of predictive genes can offer cancer researchers valuable insight into which genes may influence specific cancer types. When a sparse model works well with this minimal set of features, the model itself becomes a compact summary of the key information in the data. Additionally, sparse models reduce data storage and future data collection can be limited to the most relevant inputs.

This process of identifying a minimal yet sufficient set of features resonates with the principles of compressed sensing, a field that has shown how sparse signals can be accurately recovered from limited information.

\subsection{Background on compressed sensing}

Given an $n\times p$ input matrix $X$ with $p>n$ and an observation vector ${\bf y}\in{\mathbb R^n}$, compressed sensing assumes the linear association ${\bf y}=X \boldsymbol{\beta}^*$ holds and that the unknown vector ${\boldsymbol \beta}^*$ is $s$-sparse, that is, only $s$ of its $p$ entries are nonzero. To retrieve ${\boldsymbol \beta}^*$ with less equations than parameters ($n<p$), compress sensing aims at minimizing a cost function defined in terms of sparsity: 
\begin{equation}\label{eq:CS0}
	\min_{{\boldsymbol \beta} \in{\mathbb R}^p} \| {\boldsymbol \beta} \|_0 \quad {\rm s.t.}  \quad{\bf y}=X {\boldsymbol \beta},
\end{equation}
where $\| {\boldsymbol \beta} \|_0=s$ counts the number $s$ of nonzero entries of $ {\boldsymbol \beta}$. Solving this discrete and high-dimensional optimization problem is computationally intractable. Instead, Basis Pursuit \citep{CDS99} calculates $ \hat {\boldsymbol \beta}^{\rm BP}$ solution to the continuous optimization problem
\begin{equation}\label{eq:CS1}
	\min_{{\boldsymbol \beta} \in{\mathbb R}^p} \| {\boldsymbol \beta} \|_1 \quad {\rm s.t.}  \quad {\bf y}=X {\boldsymbol \beta},
\end{equation}
where $\|\boldsymbol{\beta} \|_1$ serves as a convex approximation of the sparsity of $\boldsymbol \beta$. For certain random matrices $X$, \citet{DonohoDL06} and \citet{Candes:Romberg:2006} proved that \eqref{eq:CS0} and \eqref{eq:CS1} can lead to $ \hat {\boldsymbol \beta}^{\rm BP}= {\boldsymbol \beta}^*$, and that a \textit{phase transition} occurs: the probability of retrieving ${\boldsymbol \beta}^*$ with Basis Pursuit is high when $s$ or $p$ are small, but then quickly drops to zero when the number $p$ of columns of $X$ or the sparsity index $s$ grows too large compared to $n$.

In real life, the observation vector ${\bf y}$ is often measured with  errors. The standard linear model assumes ${\bf y}=c{\bf 1}+X {\boldsymbol \beta}^*+{\bf e}$, where $c$ is the intercept and ${\bf e}\sim{\mathcal N}({\bf 0}, \sigma^2 I_n) $ is a Gaussian noise vector. Basis Pursuit Denoising extends Basis Pursuit~\eqref{eq:CS0} to the noisy scenario, also known as the LASSO~\citep{Tibs:regr:1996}, by solving
\begin{equation}\label{eq:LASSO}
	\min_{c\in{\mathbb R}, {\boldsymbol \beta} \in{\mathbb R}^p} \|{\bf y}-c{\bf 1}-X {\boldsymbol \beta}\|_2^q + \lambda  \| {\boldsymbol \beta} \|_\nu,
\end{equation}
for $q=2$ and $\nu=1$, where $\lambda>0$ controls the amount of sparsity in the solution. The aim here is no longer to retrieve exactly ${\boldsymbol \beta}^*$, but rather accurate identification of its support ${\cal S}^*:=\left\{j\in \{1,2,\ldots,p \}: \beta_j^*\neq 0 \right\}$, the set of  indices of the $s^*=|{\cal S}^*|$ needles hidden in the middle of a haystack of $p$ inputs. So letting $\hat {\cal S}_\lambda:=\left\{j\in \{1,2,\ldots,p \}: \hat  \beta_{\lambda,j}\neq 0 \right\}$ where $\hat {\boldsymbol \beta}_\lambda$ is a solution to~\eqref{eq:LASSO}, the natural question becomes: for a well chosen $\lambda$, is there a phase transition in the probability of exact support recovery
\begin{equation} \label{eq:PESR}
	{\rm PESR}:={\mathbb P}(\hat {\cal S}_\lambda={\cal S}^*) ?
\end{equation}
And if so, how can one select $\lambda$ from the data with a similar property?
Empirically,  \citet{Giacoetal17} and \citet{DesclouxSardy2018} observe this phase transition for a prescribed $\lambda$, and
\citet{6034731} mathematically derive promising results by proving a phase transition in the mean squared error.

\subsection{Related work on sparse learners}

By modifying \eqref{eq:LASSO}, other generalizations of Basis Pursuit Denoising include square root-LASSO~\citep{BCW11} for $q=1$, group LASSO for $\nu=2$ \citep{Yuan:Lin:mode:2006} and LASSO-zero~\citep{DesclouxSardy2018}. Under some regularity conditions, such approaches retrieve  ${\cal S}^*$ despite the noise and the high-dimensionality~\citep{BuhlGeer11}. Yet, \citet{earlyLASSO} proved that false discoveries occur early on the LASSO path, causing the phase transition to drop to zero early. To push the phase transition  further, \citet{RickChartrand_2007} and \citet{ChartrandYin08} studied and provided algorithms that replace the $\ell_1$ sparsity inducing penalty by the continuous, but non-convex, $\ell_\nu$ penalty with $\nu<1$. The idea it to get closer to the $\ell_0$ penalty than with the  $\ell_1$ penalty by letting $\nu$ get close to zero but strictly positive to deal with a continuous penalty. On the other hand, model selection based on information criteria, such as AIC \citep{AkaikeIEEE73} or BIC \citep{Schw:esti:1978}, seek sparsity using a discrete  $\ell_0$ regularizer. 

While these techniques are well established for linear models, they typically rely on cross-validation or on a validation set to select regularization parameters, and hence require fitting the model multiple times. For ANNs this becomes impractical since training even a single model can be computationally expensive. Some regularization methods have already been developed to enforce sparsity to the weights of ANNs. For example, dropout, although not explicitly a sparsity method, leaves out neurons during training, encouraging redundancy reduction and implicit feature selection. Other methods extend the idea of LASSO to neural networks, such as LassoNet \citep{LASSONET}, which applies an $\ell_1$ penalty to an input-to-output residual connection and enforces a hierarchical constraint, allowing a feature to influence hidden units only if it is active in the residual path.

\subsection{Contributions and paper organization}

In this paper, we propose an alternative approach that extends model selection to ANNs, while addressing two major drawbacks common sparsity inducing methods suffer from. First, the selection of the regularization parameter is rarely addressed, and when it is, the selection is geared towards predictive performances and not feature identification. Second, the ability to recover the {\it right} set of features has not been quantified through the prism of a phase transition in the probability of support recovery, a concept that has, however, been thoroughly studied in compressed sensing and that seem natural to extend.

Specifically, we develop an automatic feature selection method for simultaneous feature extraction and generalization that is applicable to both linear models and shallow or deep ANNs. It requires only a single, explicitly defined regularization parameter $\lambda$, eliminating the need for  cross-validation or a validation set. Our approach is straightforward to implement and easy to interpret. For ease of exposition, we present our novel method in the context of regression and classification, noting that the ideas can be ported beyond. 

This paper is organized as follows. Section~\ref{sct:proposal} describes the method by first reviewing ANNs and penalized regularization in Section~\ref{sct:featureselANN}, by proposing a definition of a sparsity inducing penalty in Section~\ref{subsct:SIP}, and by deriving  in Section~\ref{subsct:lambdapurenoise} a validation-free choice of $\lambda$ for a class of penalties. Section~\ref{sct:opti} discusses optimization issues.
Section~\ref{sct:phase_transition} empirically evaluates our method, performing a phase transition analysis on simulated data.
Section~\ref{sct:real_world} quantifies the trade--off between generalization and number of selected inputs based on fourteen real data sets.
Section~\ref{sct:conclusions} draws some general conclusions. Mathematical derivations are postponed to the Appendix.

\section{Learning how to control sparsity}\label{sct:proposal}

\subsection{Setting and notation}\label{sct:featureselANN}

To illustrate how to train a learner for sparsity, we consider the well known linear models and the artificial neural network (ANN).
A fully connected ANN, also known as a multilayer perceptron (MLP), with $L$ layers is a class of functions of the form 
\begin{equation}
    \mu_{{\boldsymbol \theta}}({\bf x})=  S_L \circ \ldots \circ S_1( {\bf x}),
\end{equation}
where ${\boldsymbol \theta}$ are the parameters indexing the ANN. An MLP is a nonlinear function that maps $\mathbb R^{p_1}$ into $\mathbb R^m$, where $p_1:=p$ the length of the input vector $\mathbf{x}$ and $m$ is the output dimension. In regression tasks, the target $y\in\mathbb R$ is a scalar (i.e. $m=1$), while in $m$-class classification tasks, ${\bf y}\in{\mathbb R}^m$ is typically a one-hot vector. 

Letting $p_2\ldots, p_L$ be the number of neurons in layer $1$ to $L-1$, the nonlinear functions $S_l({\bf u})=\sigma({\bf b}_l + W_l\bf u)$ maps the $p_l\times 1$ vector ${\bf u}$ into a $p_{l+1}\times 1$ latent vector by applying an affine transformation followed by a nonlinear activation function $\sigma$ componentwise, for each layer $l \in \{1,\ldots, L-1\}$. For the last layer, $p_{L+1}=m$ to match the output dimension of ${\bf y}\in \mathbb R^m$, so the last function is $S_L({\bf u})=\mathbf{c}+W_L\bf u$, where $W_L$ is $m \times p_L$ and $\mathbf{c}=\mathbf{b}_L$ is the intercept. No nonlinearity is applied at the output layer, since any required output transformation (such as a softmax or sigmoid) can be incorporated within the loss function rather than the network itself.

With ANN, relevant features can be identified as the columns of the weight matrix $W_1$ that are non-zero, as these are the only ones directly connected to the input layer. In order for this to work, we need to modify the nonlinear functions slightly. Specifically, for layers $l= 2, \ldots, L-1$, $S_l({\bf u}) = \sigma({\bf b}_l + W_l^\circ {\bf u})$ and $S_L(\mathbf{u})=\mathbf{c} + W_L^\circ {\bf u}$. Here $W_l^\circ$ denotes a $\ell_2$ rowwise normalized version of $W_l$. This normalization ensures that small weights in the first layer cannot be arbitrarily compensated by large weights in deeper layers, thereby preserving the impact of any regularization or penalty applied to $W_1$. Moreover, we require the activation function $\sigma$ to be unbounded (above) and have a bounded derivative. The bounded derivative assumption is satisfied by most commonly used activations. The requirement that $\sigma$ be unbounded allows to map all of ${\mathbb R}^m$. Common examples of activation functions satisfying both conditions include the ReLU, leaky ReLU and Softplus functions. We call an MLP with such modifications on its weights and such conditions on the activation function  a \textit{feature selection compatible MLP}.

The parameters indexing the neural network can be decomposed as 
\begin{equation} \label{eq:theta12}
	{\boldsymbol \theta}=\bigl(W_1,(W_2^\circ, \ldots, W_L^\circ, {\bf b}_1, \ldots, {\bf b}_{L-1}, {\bf c})\bigr)=\left(\boldsymbol{\theta}^{(1)}, \boldsymbol{\theta}^{(2)}\right)
\end{equation}
for a total of $\gamma=\sum_{k=1}^L p_{k+1}(p_k+1)$ parameters. Note that a linear learner $\mu_{(\boldsymbol{\theta}^{(1)}, \boldsymbol{\theta}^{(2)})}(\mathbf{x}) = c + \boldsymbol{\beta}^{\top} \mathbf{x}$, where $(\boldsymbol{\theta}^{(1)}, \boldsymbol{\theta}^{(2)}) = (\boldsymbol{\beta}, c)$, is a specific case of a such an MLP. This decomposition enables a natural extension of LASSO regularization to neural networks:
\begin{equation}
\min_{{\boldsymbol \theta}^{(1)} \in{\mathbb R}^{\gamma_1}, {\boldsymbol \theta}^{(2)} \in{\mathbb R}^{\gamma_2}} \|{\cal Y}-\mu_{({\boldsymbol \theta}^{(1)}, {\boldsymbol \theta}^{(2)})}({\cal X})\|_2^2+\lambda\|\boldsymbol{\theta}^{(1)}\|_1,
\end{equation}
where $( {\cal X}, {\cal Y})$ is the training set, a collection of pairs of inputs--outputs $( {\bf x}_i,{\bf y}_i)_{i=1,\ldots,n}$.
More generally, we may write the regularized training problem as
\begin{equation}\label{eq:CS2}
	\min_{{\boldsymbol \theta}^{(1)} \in{\mathbb R}^{\gamma_1}, {\boldsymbol \theta}^{(2)} \in{\mathbb R}^{\gamma_2}} {\cal L}_n(\mu_{({\boldsymbol \theta}^{(1)}, {\boldsymbol \theta}^{(2)})}; {\cal Y}, {\cal X}) +  P({\boldsymbol \theta}^{(1)}, \lambda).
\end{equation}
where ${\cal L}_n$ is a data-dependent goodness-of-fit measure (e.g., negative log-likelihood function, cross-entropy) also called the loss function, $P$ is a sparsity inducing penalty,  ${\cal X} \in \mathbb{R}^{n \times p}$ denotes the matrix of input covariates and ${\cal Y}\in \mathbb{R}^{mn}$ is the corresponding response vector. The complete expression is referred to as the cost function. Not all coefficients are penalized:  this allows for instance  the intercept in linear regression to not be penalized. Also, since with ANN the loss term in~\eqref{eq:CS2} is not convex, there is no longer an incentive to employ LASSO’s convex penalty $P({\boldsymbol \theta}^{(1)}, \lambda) = \lambda\|{\boldsymbol \theta}^{(1)}\|_1$. Instead, one may consider non-convex continuous penalties that better approximate $P({\boldsymbol \theta}^{(1)}, \lambda) = \lambda\|{\boldsymbol \theta}^{(1)}\|_0$.

\subsection{Sparsity inducing penalties}\label{subsct:SIP}

Our methodology relies on the well known key concept of a sparsity inducing penalty, of which we give our precise definition by a local behavior at the origin.

\begin{definition}[Sparsity-inducing penalty]\label{def:sparsindupen}
The penalty function $P$ in the optimization problem~\eqref{eq:CS2} is said to be sparsity-inducing if there exists a finite value $\lambda > 0$ such that all $(\hat{\boldsymbol{\theta}}^{(1)}, \hat{\boldsymbol{\theta}}^{(2)}) = (\mathbf{0}, \hat{\boldsymbol{\theta}}^{(2)})$ are local minimizers.
\end{definition}

It is well known that the ridge penalty \citep{ridgeHK}, $P({\boldsymbol \theta}^{(1)}, \lambda)=\lambda\|{\boldsymbol \theta}^{(1)}\|_2^2$, does not qualify as a sparsity-inducing penalty. While it shrinks coefficients towards zero, it does not eliminate any features from ${\boldsymbol \theta}^{(1)}$. In contrast, the $\ell_1$ LASSO penalty, $P({\boldsymbol \theta}^{(1)}, \lambda) = \lambda\|{\boldsymbol \theta}^{(1)}\|_1$, promotes sparse solutions and satisfies the definition. The $\ell_0$ penalty, $P({\boldsymbol \theta}^{(1)}, \lambda) = \lambda\|{\boldsymbol \theta}^{(1)}\|_0$, also meets this criterion by explicitly penalizing the number of non-zero elements.

Beyond sparsity, desirable penalty functions should ideally result in estimators that are nearly unbiased for large true parameter values to avoid unnecessary modeling bias, and that are continuous to avoid instability in model prediction. While the $\ell_0$ penalty is theoretically optimal in terms of bias reduction, it is discontinuous and leads to an NP-hard optimization problem, making it computationally infeasible in practice. So \citet{SCAD} and \citet{MCP} proposed the Smoothly Clipped Absolute Deviation (SCAD) and the Minimax Concave Penalty (MCP), respectively. These functions are specifically designed to induce sparsity while offering less bias for large coefficients. 

In the same way, we propose a class of sparsity-inducing penalties spanning a continuum between $\ell_0$ and $\ell_1$: the closer to $\ell_0$, the further from convexity, but the less shrinkage of the relevant coefficients to zero for better generalization. We consider the penalty function 
\begin{equation}\label{eq:Pnu}
	P_\nu({\boldsymbol \theta}, \lambda)=  \lambda\sum_{j=1}^{p}\rho_\nu(\theta_j) \quad {\rm with} \quad \rho_\nu(\theta) =\frac{|\theta|}{1+|\theta|^{1- \nu}},
\end{equation}
where the entries of ${\boldsymbol \theta}$ are entered componentwise in the penalty. For $\nu=1$ it amounts to LASSO's $\ell_1$ convex penalty (albeit a factor $1/2$) and for $\nu\rightarrow 0$ it tends to the $\ell_0$ discrete penalty for $|\theta|$ large. 
The reason for the introduction of this new penalty is threefold. First, unlike SCAD and MCP, which require piecewise definitions and per‐coordinate branching, $\rho_\nu(\theta)$ is one closed‐form expression. This yields fully vectorized GPU kernels with no conditional masks that can be computationally expensive when $\boldsymbol{\theta}^{(1)}$ has many entries, as it is the case for ANNs. Second, SCAD’s parameter $a>2$ or MCP’s $\gamma>0$ affect non‐convexity in a non‐transparent way. On the contrary with \eqref{eq:Pnu},  $\nu=1$ recovers $\ell_1$, and $\nu\to0$ approaches $\ell_0$, so a simple grid or annealing schedule $\nu\in\{0.9,0.7,\dots,0.1\}$ directly interpolates between them. Third, SCAD and MCP introduce kinks where their derivative jumps; this can slow or destabilize training. In contrast, $\rho_\nu(\theta)$ is continuously differentiable for all $\theta\neq 0$, ensuring stable, predictable backpropagation. Regardless of what approximation of $\ell_0$ is used, the penalty is isotropic, so the columns of the input matrix $\mathcal{X}$ must be rescaled, for instance by dividing them by their respective standard deviations. Throughout the paper, we  assume that this standardization has been performed.

The following theorem illustrates in the univariate case with the $\ell_2$-loss how the new penalty compares to the hard- and soft-thresholding functions  that provide a closed form expression to the  $\ell_0$- and $\ell_1$-penalized least squares, respectively \citep{Dono94b}. Considering the univariate case is important because the optimization scheme used to solve~\eqref{eq:CS2} in Section~\eqref{sct:opti} involves using the univariate solutions iteratively until convergence to a stationary point.

\begin{figure}
\centerline{\includegraphics[width=\textwidth]{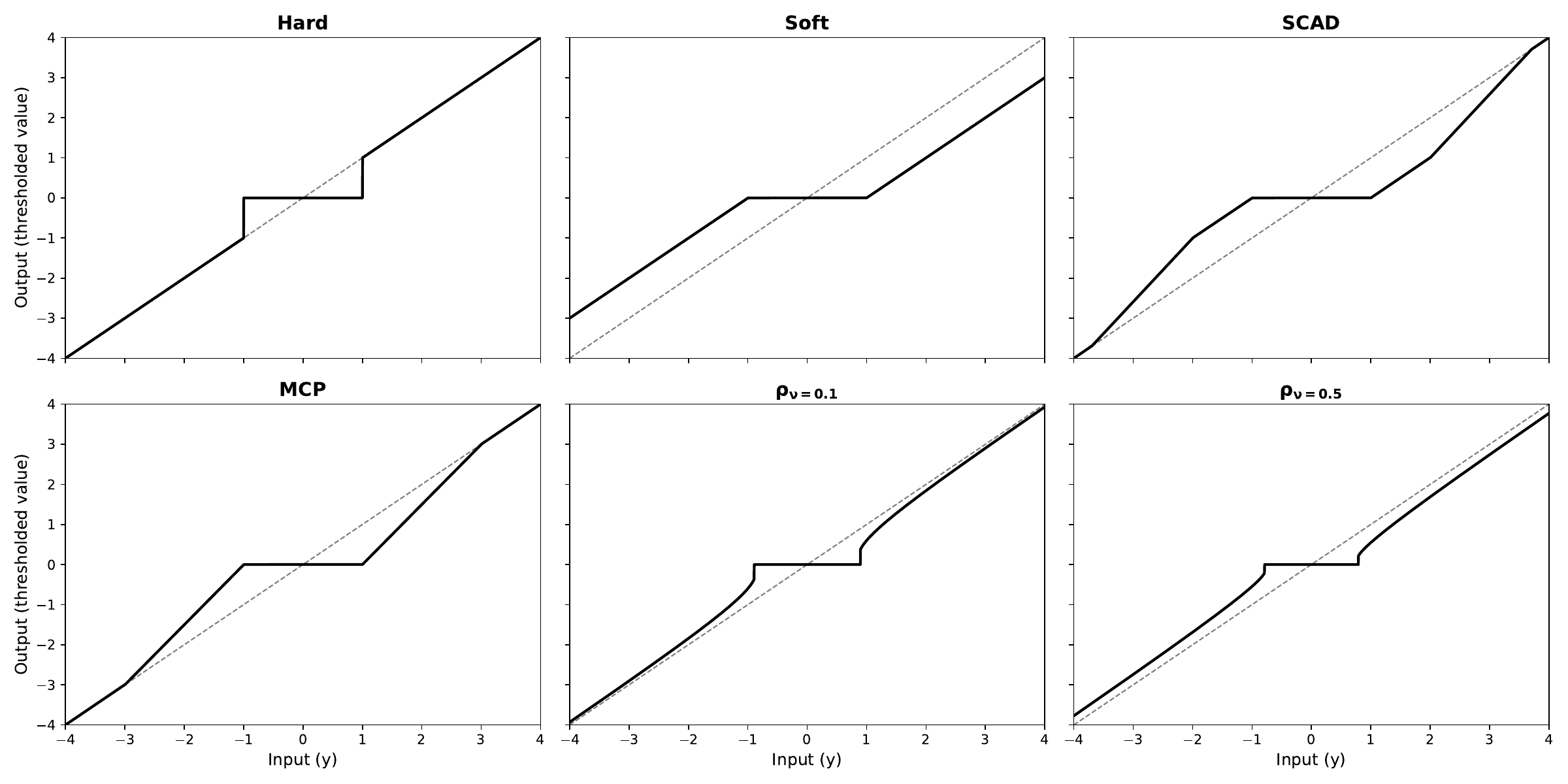}}
\caption{Plot of Thresholding Functions for the Hard $\ell_0$, the Soft $\ell_1$, the SCAD (with $a=3.7$), the MCP (with $\gamma=3.0$), the ``Harder`` $\rho_{\nu=0.1}$ and $\rho_{\nu=0.5}$ Thresholding Functions with $\lambda=1$. Dotted line: the identity function.} 
\label{fig:thresholdingfct}
\end{figure}

\begin{theorem}[Univariate thresholding]
	\label{th:rho1D}
	Given fixed $\lambda>0$ and $\nu\in(0,1]$
	, the solution $\hat \theta$ to
	\begin{equation} \label{eq:1D}
		\min_{\theta \in {\mathbb R}} \frac{1}{2} (y-\theta)^2+ \lambda \rho_\nu(\theta)
	\end{equation}
	as a function of $y$ is a thresholding function $\eta_\lambda(y;\nu ):=\hat \theta$. This thresholding function is discontinuous at its threshold $ \varphi(\lambda,\nu)$ with a jump $ \kappa(\lambda,\nu)$ solutions to the system 
	\begin{equation}\label{eq:varphikappa}
		\left \{
		\begin{array}{l}
			\kappa^{2-\nu}+2\kappa + \kappa^\nu + 2 \lambda(\nu-1)=0 \\
			\varphi=\kappa/2+  \lambda  \frac{1}{1+\kappa^{1- \nu}}
		\end{array}
		\right. .
	\end{equation}
\end{theorem}
Figure~\eqref{fig:thresholdingfct} plots the thresholding function for $\lambda=1$ along with other popular thresholding functions. The parameter $\nu=0.1$ being close to zero, one sees that $\eta_\lambda(y;0.1 )$ approximates the hard thresholding function $\eta_{\varphi_0}(y;{\rm hard})=y \cdot 1_{\{|y|>\varphi_0\}}$ with $\varphi_0=\varphi(1,0.1)$. Since we aim at approaching the hard thresholding function in a continuous way by letting $\nu$ be small, we call the corresponding method {\it HarderLASSO}. It also bears its name from the fact that the corresponding optimization is harder to solve when $\nu$ gets close to zero; so we restrict nearing $\ell_0$ by choosing $\nu=0.1$ in the sequel. One expects a better phase transition for the probability of recovering the relevant features with the new penalty when $\nu=0.1$ then when $\nu=1$. 

\subsection{Choice of $\lambda$ under pure noise}\label{subsct:lambdapurenoise}
The regularization parameter $\lambda$ plays a critical role in the optimization problem~\eqref{eq:CS2} that defines the learner. If $\lambda$ is too large, the penalty term dominates the loss and forces $\hat{\boldsymbol \theta}^{(1)} = \mathbf{0}$. If $\lambda$ is too small, sparsity is insufficiently enforced, and irrelevant features may be selected. Therefore, careful tuning of $\lambda$ is essential for successful feature selection and generalization. 

The standard approach to selecting $\lambda$ is through cross-validation. This method is often suboptimal however: it is computationally intensive, requires splitting the dataset, and tends to select overly complex models with limited interpretability. Another similar option is through a validation set, when it is available. Both validation-based methods are geared towards generalization and tend to select $\lambda$ too small ruining a clear phase transition.

An alternative is to calibrate $\lambda$ under the pure noise model (that is, ${\boldsymbol{\theta}}^{(1)}={\bf 0}$) such that, with high probability $1-\alpha$ for some small prescribed $\alpha$, the cost function~\eqref{eq:CS2} has a local minimum with $\hat{\boldsymbol{\theta}}^{(1)}={\bf 0}$. This calibration of $\lambda$ has seen some success for Gaussian linear regression \citep{Dono94b, BCW11} and is supported by the LASSO theory \citep{BuhlGeer11}. The Quantile Universal Threshold (QUT) generalizes this concept~\citep{Giacoetal17}.

Achieving the desired probabilistic property requires $\lambda$ to be sufficiently large to set all penalized parameters to zero. The question becomes: how large? 
Recalling Definition~\ref{def:sparsindupen} of a sparsity inducing penalty $P$, answering this question is straightforward if one knows the {\it finite} value of $\lambda$ that garantees local minima at the origin. Indeed, suppose there exists a function $\lambda_0({\cal X}, {\cal Y})$ of the data that provides the smallest $\lambda$ that creates (local) solutions to~\eqref{eq:CS2} with
$\hat{\boldsymbol{\theta}}_\lambda^{(1)}={\bf 0}$ (Appendix~\ref{app:zero-thresh} proves existence and derives formula for this function). Then, letting the random variable $\Lambda=\lambda_0({\cal X}, Y_0)$, where $Y_0$ are random data outputs simulated under the pure noise assumption, one sees that $ {\mathbb P}(\hat{\boldsymbol{\theta}}^{(1)}_\lambda={\bf 0})={\mathbb P}\left(\lambda \geq \lambda_0({\cal X}, Y_0)\right)$. So if one sets $\lambda$ to $\lambda^{\rm QUT}_\alpha= F^{-1}_\Lambda(1-\alpha)$, then $ {\mathbb P}(\hat{\boldsymbol{\theta}}^{(1)}_\lambda={\bf 0})=1-\alpha$. This probabilistic property leads to the definition of the quantile universal threshold.

\begin{definition}[Quantile universal threshold] \label{def:QUT}
  Given some small probability  $\alpha$, the quantile universal threshold (QUT) for~\eqref{eq:CS2}  is the upper $\alpha$-quantile
  $\lambda_\alpha^{\mathrm{QUT}} = F^{-1}_\Lambda(1-\alpha)$
  of the statistic $\Lambda = \lambda_0({\cal X}, Y_0)$, where $\lambda_0({\cal X}, {\cal Y})$ is the zero-thresholding function of~\eqref{eq:CS2} and
  $Y_0$ is the random vector when no input has any predictive power.
\end{definition}
When the data do not come from the null model but are sufficiently close to it (that is, when the number $s$ of significant variables is small), this same threshold $\lambda_\alpha^{\mathrm{QUT}}$ keeps high the probability of exact support recovery, which empirically points  to  a phase transition in the probability of correctly finding relevant variables. In other words, QUT is explicitly calibrated for accurate support recovery rather than for minimizing prediction error. As for the value $\alpha$, it amounts to the false discovery rate under the pure noise setting and is typically chosen to $\alpha=0.05$.
 
The derivation of $\lambda_\alpha^{\mathrm{QUT}}$ relies on the so-called zero-thresholding function, which has the following closed form expression for regression and classification (see Appendix~\ref{app:zero-thresh}).
 
\begin{lemma}[Zero-thresholding function for MLPs]\label{lem:lambda_regclass}
  For a training set $({\cal X},{\cal Y})$, the zero-thresholding function for the optimization problem~\eqref{eq:CS2} with the penalty $P_\nu$ \eqref{eq:Pnu}
   and a feature selection compatible MLP with $L$ layers is:
     \begin{itemize}
      \item For regression with square root least squares loss:
  \begin{equation}
  	\lambda_0( {\cal X}, {\cal Y}) =
  \kappa^{L-1}\pi_L\frac{\|{\cal X}^{\mathrm{T}}({\cal Y}-\bar{{\cal Y}}
  \mathbf{1}_n) \|_\infty}{ \|{\cal Y}-\bar{{\cal Y}}  \mathbf{1}_n
  \|_2};
  \end{equation}
      \item For classification with cross-entropy loss:
  \begin{equation}
  	\lambda_0({\cal X}, {\cal Y}) =
  \kappa^{L-1}\pi_L\|{\cal X}^{\mathrm{T}}({\cal Y}-\bar{{\cal Y}}
  \mathbf{1}_n) \|_\infty,
  \end{equation}
  \end{itemize}
  where $\kappa = \sup_t |\sigma'(t)|$ and $\pi_L = \sqrt{\prod_{j=3}^L
  p_j}$ for $L \geq 3$, $\pi_2 = \pi_1 = 1$ and $\|A\|_{\infty}=$ $\max _{j=1, \ldots, p} \sum_{i=1}^k\left|a_{j i}\right|$ for a $p \times k$ matrix $A$.
\end{lemma}
For Gaussian regression, the null distribution is ${\bf Y}_0 \sim \mathcal{N}(c \mathbf{1}, \xi^2 I_n)$. A key advantage  is that $\Lambda$ depends only on the mean-centered responses ${\bf Y}_0 - \bar{\bf Y}_0\mathbf{1}_n$ which do not depend on $c$,  and on the ratio of two quantities proportional to $\xi$; so $\Lambda$ is location- and scale-invariant. Consequently, $\Lambda$ is a pivotal random variable in the Gaussian case, meaning that knowledge of $c$ and $\xi$ is not required to derive $\lambda_\alpha^{\mathrm{QUT}}$. This well-known fact  \citep{BCW11} motivates the use of $\mathcal{L}_n=\left\|{\cal Y}-\mu_{\boldsymbol{\theta}}({\cal X})\right\|_2$ rather than $\mathcal{L}_n=\left\|{\cal Y}-\mu_{\boldsymbol{\theta}}({\cal X})\right\|_2^2$.
For classification, we assume the null distribution is $Y_0 \sim \operatorname{Multinomial}(n, \mathbf{p} = \hat{\mathbf{p}})$, where $\hat{\mathbf{p}}$ are the observed class proportions in the training set.

The upper quantile of $\Lambda$ can be easily estimated by Monte Carlo simulation. This approach eliminates the need for cross-validation or a separate validation set, making it particularly attractive for neural networks where training is computationally expensive.

\section{Optimization}\label{sct:opti}

Focusing on solving~\eqref{eq:CS2} with the regularization parameter $\lambda$ and sparsity-inducing penalty~$P$ recommended  in the previous sections, our learner is defined as a solution to 
\begin{equation}\label{eq:CS3}
	\min_{{\boldsymbol \theta}^{(1)} \in{\mathbb R}^{\gamma_1}, {\boldsymbol \theta}^{(2)} \in{\mathbb R}^{\gamma_2}} {\cal L}_n(\mu_{({\boldsymbol \theta}^{(1)}, {\boldsymbol \theta}^{(2)})}; {\cal Y}, {\cal X}) +  \lambda_{\alpha}^{\rm QUT} \sum_{j=1}^p \rho_\nu(\boldsymbol{\theta}^{(1)}_j),
\end{equation}
where $\mu_{\boldsymbol{\theta}}$ is a linear function or a feature selection compatible MLP as described in Section~\ref{sct:featureselANN}, $\rho_{\nu}$ is our nonconvex penalty function given in \ref{eq:Pnu} with $\nu=0.1$ and $\lambda_{\alpha}^{\rm QUT}$ is the QUT associated to the loss function $\mathcal{L}$ for a small $\alpha$, say $\alpha=0.05$.


\subsection{Annealing schedule}
While the ultimate goal is to solve \eqref{eq:CS3} for $(\lambda, \nu)=\left(\lambda_\alpha^{\text {QUT }}, 0.1\right)$, we take the conservative approach of solving for a sequence of $(\lambda, \nu)$'s to prevent falling in a poor local minimum, in the spirit of simulated annealing. Namely, we propose solving \eqref{eq:CS3} using $\lambda_i=\exp{(i-1)} /(1+\exp{(i-1)}) \lambda_\alpha^{\rm QUT }$ for $i \in\{0, \ldots, 5,+\infty\}$ while $\nu$ takes on values from the set $\{0.9,0.7,0.4,0.3,0.2,0.1\}$. Taking as initial parameter values the solution corresponding to the previous $\lambda_i$ leads to a sequence of sparser approximating solutions until solving for $\lambda_\alpha^{\rm QUT }$ at the last step. This gradual sequence helps mitigate the risk of falling into a poor local minima due to overly aggressive thresholding early in the process.

\subsection{Gradient and proximal updates}
The first phases of the annealing schedule with successive $(\lambda_i, \nu_i)$ for $i=0,\ldots,5$ aim at slowly drifting towards a good local minimum to~\eqref{eq:CS3}, maybe even global, without seeking a sparse solution; we do so with a standard gradient descent optimizer (e.g., the Adam optimizer with an initial learning rate of $0.01$, while keeping the remaining optimizer parameters at their default values), neglecting the non-differentiability of the penalty term. Because the solution for  $(\lambda_i, \nu_i)$ is only used as a warm start for solving the next optimization problem with $(\lambda_{i+1}, \nu_{i+1})$, we use a large convergence threshold since an approximate solution is sufficient. 

The final phase (i.e., with $(\lambda_i, \nu_i)=(\lambda_\alpha^{\rm QUT}, 0.1)$) requires precision and special care of the non-differentiability issue to get a truly sparse solution. To that aim, we use ISTA \citep{FISTA09}, an algorithm specifically designed for dealing with a sparsity inducing penalty.
Applying an ISTA step to minimize an optimization problem of the form $f(\theta)+g(\theta)$ entails moving from the current iterate $\boldsymbol{\theta}_{(k)}$ to $\boldsymbol{\theta}_{(k+1)}$ by iteratively solving
\begin{equation} \label{eq:FISTAstep}
	\min _{\boldsymbol{\theta} \in \mathbb{R}^\gamma} \frac{1}{2 \delta_{(k)}}\left\|\left(\boldsymbol{\theta}_{(k)}-\delta_{(k)} \nabla f\left(\boldsymbol{\theta}_{(k)}\right)\right)-\boldsymbol{\theta}\right\|_2^2+g\left(\boldsymbol{\theta}_{(k)}\right),
\end{equation}
 where $\delta_{(k)}$ is a step size (similar to a learning rate). Here $g$ plays the role of the nondifferentiable sparsity inducing penalty. The multivariate optimization~\eqref{eq:FISTAstep} is separable and amounts to solving $\gamma=\sum_{k=1}^l p_{k+1}(p_k+1)$ univariate problems, either least squares problems for $\boldsymbol{\theta}^{(2)}$ or the univariate problem of Theorem \ref{th:rho1D} for $\boldsymbol{\theta}^{(1)}$. 
So the penalized parameters are updated using ISTA and a line search is used to determine the step size. Each training phase is concluded when the relative improvement in the cost falls below a threshold, large for $i=0,\ldots,5$ and small for the final phase.

\subsection{Final fit on sparse estimated model}
After completing all training phases, we reduce the matrix $\hat W_1$ to $\hat W_1 \in \mathbb{R}^{\tilde{p}_2 \times \hat s}$, where $\hat s$ is the number of selected features by the model. This is done by removing any neurons that consist entirely of zeros, resulting from the regularization applied to the matrix $W_1$. This adjustment impacts the subsequent layer's weight matrix, so that $\hat W_2^{\circ}$ is now reshaped to $\hat W_2 ^\circ\in \mathbb{R}^{p_3 \times \tilde{p}_2}$, where $\tilde{p}_2$ reflects the reduced number of active neurons. A final training phase without regularization is then applied to the reduced model.

\subsection{Computational Complexity}
Except for the first step of the annealing schedule for $i=0$, all the other steps (including the last FISTA step) use the warm start, meaning that the initial values of the parameters are not far from the solution since the successive $(\lambda_i,\nu_i)$ are close to one another.  So convergence is quickly reached. The final unpenalized refit is likewise quick, since it begins from a well-aligned, sparsified network.

\section{Empirical phase transition analysis}\label{sct:phase_transition}
We now empirically validate our methodology through extensive simulations.
We examine the performance of HarderLASSO on both linear models and artificial neural networks by comparing it to various state-of-the-art methods on regression tasks that consist of predicting a scalar output $y$ from an input vector $\mathbf{x}$. We evaluate performance through the lens of a phase transition in the probability~\eqref{eq:PESR} of retrieving the relevant features and only them. To investigate the ability of learners to select the correct features and generalize effectively, we consider $s$-sparse regression problems in the sense that out of the $p$ total inputs, only $s$ of them carry predictive information. We define $\mathcal{S}^*=\{j\in\{1,\ldots,p\}:x_j\text{ has predictive information}\}$ as the true support and $\hat{\mathcal{S}}$ as the estimated support produced by a given method. So the model is $s$-sparse when $|\mathcal{S}^*|= s$. 

To measure the ability of a learner to find a good association $\mu(\cdot)$ between input and output, we consider four evaluation criteria:
\begin{itemize}
  \item Probability of Exact Support Recovery: $\operatorname{PESR}:=\mathbb{P}(\hat{\mathcal{S}}=\mathcal{S}^*)$, which measures the probability that the estimated support exactly matches the true support.
   \item Generalization Error: measured by the empirical $L_2$ distance between the true association $\mu^*$ and the estimated one $\hat{\mu}$ on an independent test set:
  \begin{equation*}
  \hat{L}_2=\frac{1}{n_{\text{test}}}\sum_{i=1}^{n_{\text{test}}}\left(\hat{\mu}\left(\mathbf{x}_i^{\text{test}}\right)-\mu^*\left(\mathbf{x}_i^{\text{test}}\right)\right)^2,
  \end{equation*}
  where $\{\mathbf{x}_i^{\text{test}}\}_{i=1}^{n_{\text{test}}}$ are drawn from an independent test set.

  \item False Discovery Rate: $\operatorname{FDR}:=\mathbb{E}\left(\frac{|\hat{\mathcal{S}} \setminus\mathcal{S}^*|}{|\hat{\mathcal{S}}| \vee 1}\right)$, which quantifies the expected proportion of incorrectly selected features among all selected features.
  \item True Positive Rate: $\operatorname{TPR}:=\mathbb{E}\left(\frac{|\hat{\mathcal{S}} \cap\mathcal{S}^*|}{|\mathcal{S}^*|}\right)$, which measures the expected proportion of correctly identified relevant features among all truly relevant features.
 \end{itemize}

Since the training sets are generated from a known $s$-sparse model that we define below, both the true support $\mathcal{S}^*$ and the true association $\mu^*(\cdot)$ are known. So all four criteria can be evaluated. The primary focus of our analysis is the phase transition phenomenon: we investigate how PESR behaves as a function of sparsity parameter $s$. PESR is a stringent criterion. Exact recovery is reached only when the estimated support matches the true support exactly, no missing or extra variables. So PESR measures success as a binary notion of optimality rather than near correctness, making it a highly demanding indicator that nonetheless exposes methods truly capable of achieving optimal feature selection.

\subsection{Data generation} \label{subsct:simdata}
To compute our different performance metrics and analyze the phase transition phenomenon, we generate $m$ independent datasets for each sparsity level $s$. Each data set is obtained from an $n\times p$ Gaussian input matrix $X$ (entries i.i.d.~standard normal) by randomly selecting $s$ columns to form the $n\times s$ submatrix $X_s^{(k)}$ for $k=1,\dots, m$. Given a true association function $\mu^*(\cdot): \mathbb{R}^s\to\mathbb{R}$, $m$ training sets are created according to 
\begin{equation*}
    \mathbf{y}_s^{(k)}=\mu^*(X_s^{(k)})+\mathbf{e}^{(k)} \quad k=1,\dots, m,
\end{equation*}
where $\mu^*$ acts rowwise on $X_s^{(k)}$ and $\mathbf{e}^{(k)}$ are i.i.d.~sampled from a standard Gaussian distribution. In the following, we consider linear and nonlinear $\mu^*$ associations. 

\subsection{Linear models}
For the linear simulation study, we fix the problem dimensions at $(n,p) = (70,250)$ and vary the sparsity index $s \in \{0, 1, \ldots, 25\}$ to examine the phase transition behavior. This configuration represents a challenging high-dimensional regime where $p\gg n$. The true association function is specified as 
      $\mu^*(X_s^{(k)}) = X_s^{(k)}\boldsymbol{\beta}^*$,
where $\boldsymbol{\beta}^* \in \mathbb{R}^s$ is the true coefficient vector, with each entry randomly drawn from $\{-3, -2, -1, 1, 2, 3\}$. The magnitude of these coefficients provides a signal strong enough to remain distinguishable from the noise.

Our method is evaluated against three established linear variable-selection techniques described in Table \ref{tab:linear_models}. All cross-validated methods are implemented using the ncvreg R package, with the regularization parameter $\lambda$ determined by 5-fold cross-validation. 

\begin{figure}
    \centering
    \includegraphics[width=\linewidth]{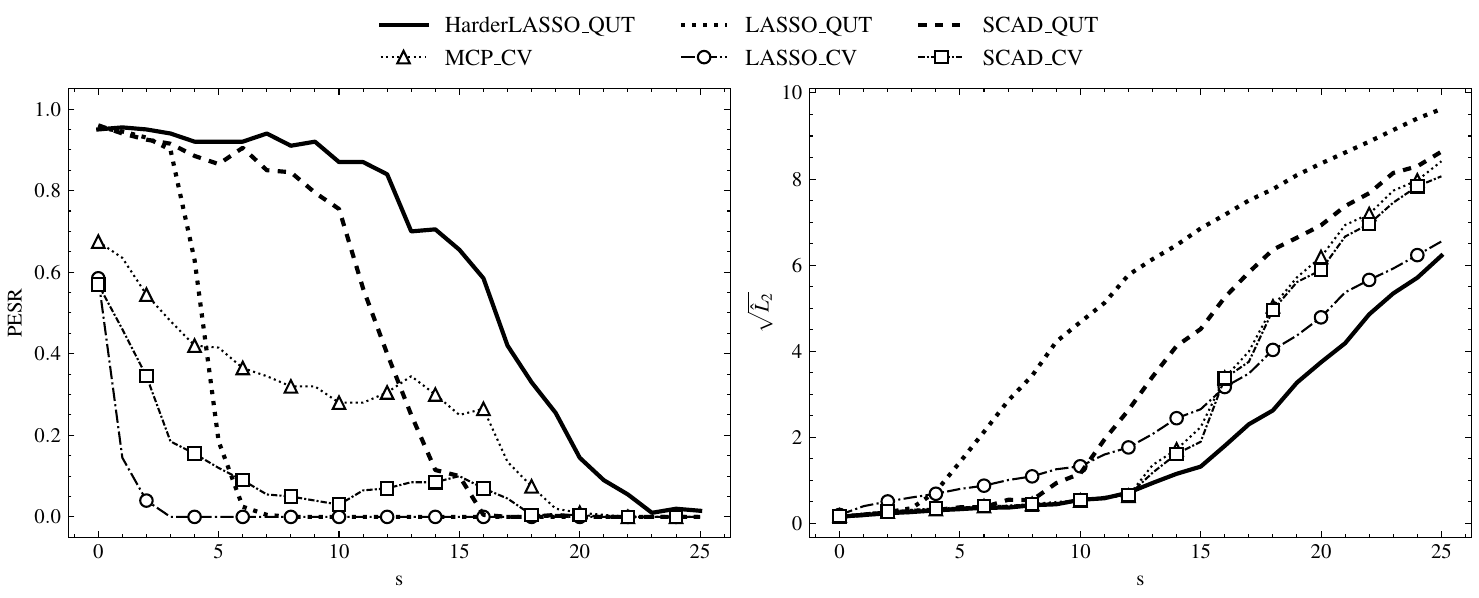}
    \caption{Results of Monte Carlo simulations ($m = 200$ runs) for the linear setting. Generalization is evaluated on an independent test set of size $n = 1000$.}
    \label{fig:linear_simu_results}
\end{figure}

Figure \ref{fig:linear_simu_results} summarizes the results of the Monte Carlo simulation with for the PESR and $\hat{L}_2$ metrics. The \texttt{HarderLASSO\_QUT} consistently outperforms all competing methods across both criteria, with the harder/non-convex penalty providing a distinct advantage over the convex \texttt{LASSO} by reducing shrinkage on non‐zero coefficients. Notably, the QUT-based learners are the only ones to have a clear phase transition, enabled by the good selection of the penalty parameter $\lambda$ with the quantile universal rule. In contrast to cross-validated methods, which are primarily optimized for predictive accuracy, QUT-based learners are explicitly tailored for feature selection. As a result, the QUT estimators uniformly surpass their cross-validated counterparts in terms of PESR. Results for the FDR and TPR can be found in the appendix.

\begin{table}[ht]
\centering
\resizebox{\textwidth}{!}{%
\begin{tabular}{ll}
\toprule
\textbf{Model} & \textbf{Description} \\
\midrule
\texttt{LASSO\_CV} & $\ell_2$-loss and $\ell_1$ penalty \citep{Tibs:regr:1996} \\
\texttt{SCAD\_CV} & $\ell_2$-loss and SCAD penalty ($\alpha=3.7$) \citep{SCAD} \\
\texttt{MCP\_CV} & $\ell_2$-loss and MCP ($\gamma=3$) \citep{MCP} \\
\texttt{LASSO\_QUT} & square-root $\ell_2$-loss and $\ell_1$ penalty with QUT regularization \\
\texttt{SCAD\_QUT} & square-root $\ell_2$-loss and SCAD ($\alpha=3.7$) with QUT regularization \\
\texttt{HarderLASSO\_QUT} & square-root $\ell_2$-loss and "harder" thresholding with QUT regularization \\
\bottomrule
\end{tabular}%
}
\caption{Summary of the linear sparse learners.}
\label{tab:linear_models}
\end{table}

\subsection{Nonlinear models with Artificial Neural Networks}
For a simple nonlinear simulation, we choose
\begin{equation*}
    \mu^*(X_s^{(k)})=\sum_{i=1}^h 10\cdot|\mathbf{x}_{2i}-\mathbf{x}_{2i-1}|,
\end{equation*}
where $\mathbf{x}_i$ is the i-th column of $X_s^{(k)}$ of Section~\ref{subsct:simdata}, and $h=s/2$ for a sparsity index~$s$ varying over the set $\{0, 2, 4, \dots, 20\}$.
This specification introduces a piecewise linear, non-monotone component via the absolute value, making variable selection and estimation more challenging than in standard linear models. To allow detection of the nonlinearity, we choose $p=50$ and a large $n=500$ so that dimensionality does not dominate the difficulty of retrieving the non-monotone dependence. 

Our method is compared against three popular nonlinear variable-selection techniques, described in Table \ref{tab:nonlinear_models}. \texttt{LassoNet} is a neural network framework that achieves feature selection by adding a skip layer and integrating feature selection directly into the objective function; we select the tuning parameters of LassoNet using the built-in 5-fold cross-validation. \texttt{Random Forest} is based on the sklearn package with a configuration of 100 trees. The \texttt{XGBoost} learner is based on the xgboost Python package with default parameter values. Given that both Random Forest and XGBoost primarily generate feature rankings rather than explicitly selecting
features, we apply the Boruta algorithm to identify the most relevant features \citep{boruta10}. After determining the selected features using Boruta, we retrain the models on the selected subset.  Both our neural network learner and \texttt{LassoNet} use one hidden layer of size $p_2 = 20$ and the ReLU activation function.

\begin{figure}
    \centering
    \includegraphics[width=\linewidth]{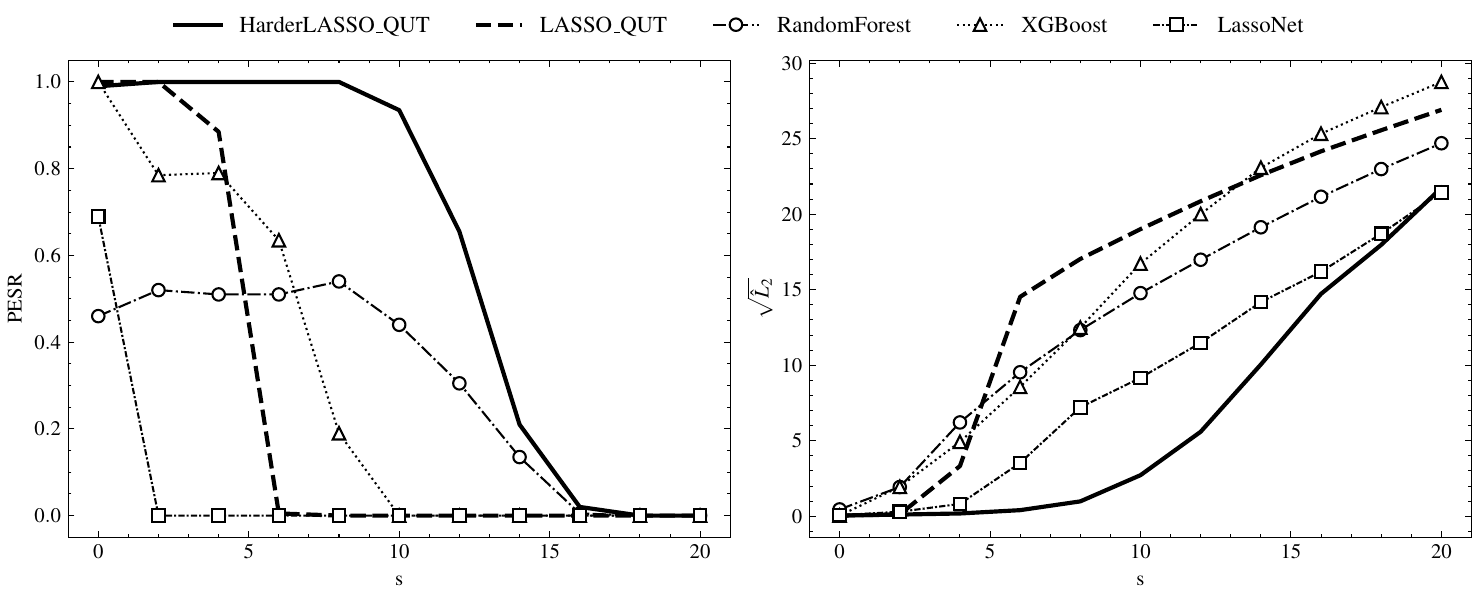}
    \caption{Results of Monte Carlo simulations ($m = 200$ runs) for the nonlinear setting. Generalization is evaluated on an independent test set of size $n = 1000$.}
    \label{fig:nonlinear_simu_results}
\end{figure}

Figure~\ref{fig:nonlinear_simu_results} summarizes the results of the nonlinear Monte Carlo simulation with $m = 200$ runs.
 One must recall that both \texttt{Random Forest} and \texttt{XGBoost} are ensemble methods that average the output of a large number of base learners, which inherently boosts stability and predictive accuracy. So this gives them a non-trivial advantage over our learner and \texttt{LassoNet}: it is like comparing CART to Random Forest. Nevertheless our learner is quite competitive: the same conclusion holds pointing to the superiority of \texttt{HarderLASSO\_QUT}. 
Results for the FDR and TRP can be found in the appendix.

\begin{table}[ht]
\centering
\resizebox{\textwidth}{!}{%
\begin{tabular}{ll}
\toprule
\textbf{Model} & \textbf{Description} \\
\midrule
\texttt{RandomForest} & Average of CART \citep{breiman2001random} \\
\texttt{XGBoost} & Extreme Gradient Boosting \citep{Chen_2016} \\
\texttt{LassoNet} & $\ell_2$-loss with penalized neural network \citep{LASSONET}. \\
\texttt{LASSO\_QUT} & square-root $\ell_2$-loss and $\ell_1$ penalty with QUT regularization \\
\texttt{HarderLASSO\_QUT} & square-root $\ell_2$-loss and "harder" thresholding with QUT regularization \\
\bottomrule
\end{tabular}%
}
\caption{Summary of the nonlinear sparse learners.}
\label{tab:nonlinear_models}
\end{table}

\subsection{Network architecture effect on the phase transition}
We finally perform an interesting experiment that reveals an additional strength of our method. 
To investigate the effect of network architectures, we consider ANNs of 1, 2 and 3 hidden layers, using widths $p_2=20$, $(p_2, p_3)=(20, 10)$ and $(p_2, p_3, p_4)=(20, 10, 5)$. For an increasing sample size $n$ and $s=4$ fixed, we empirically investigate their respective abilities to create a phase transition not only in the probability of retrieving the four features but also in unveiling the complex nonlinearity of the s-sparse function, defined as:
\begin{equation*}
    \mu^*(X_s^{(k)})=10 \big||\mathbf{x}_{2}-\mathbf{x}_{1}|-|\mathbf{x}_{4}-\mathbf{x}_{3}|\big|.
\end{equation*}
A one hidden layer ReLU network can approximate this function only if it has sufficiently many hidden neurons. Variable selection can be accurate, but the nested absolute value structure may not discovered, leading to poor generalization. In contrast, a two hidden layer network should yield optimal exact support recovery and generalization, effectively capturing the complex nonlinear associations. A three hidden layer network is also expressive enough, but the additional flexibility of adding another layer and more parameters has a cost: it  requires larger $n$ to match the performance of the two–layer model.

\begin{figure}
    \centering
    \includegraphics[width=\linewidth]{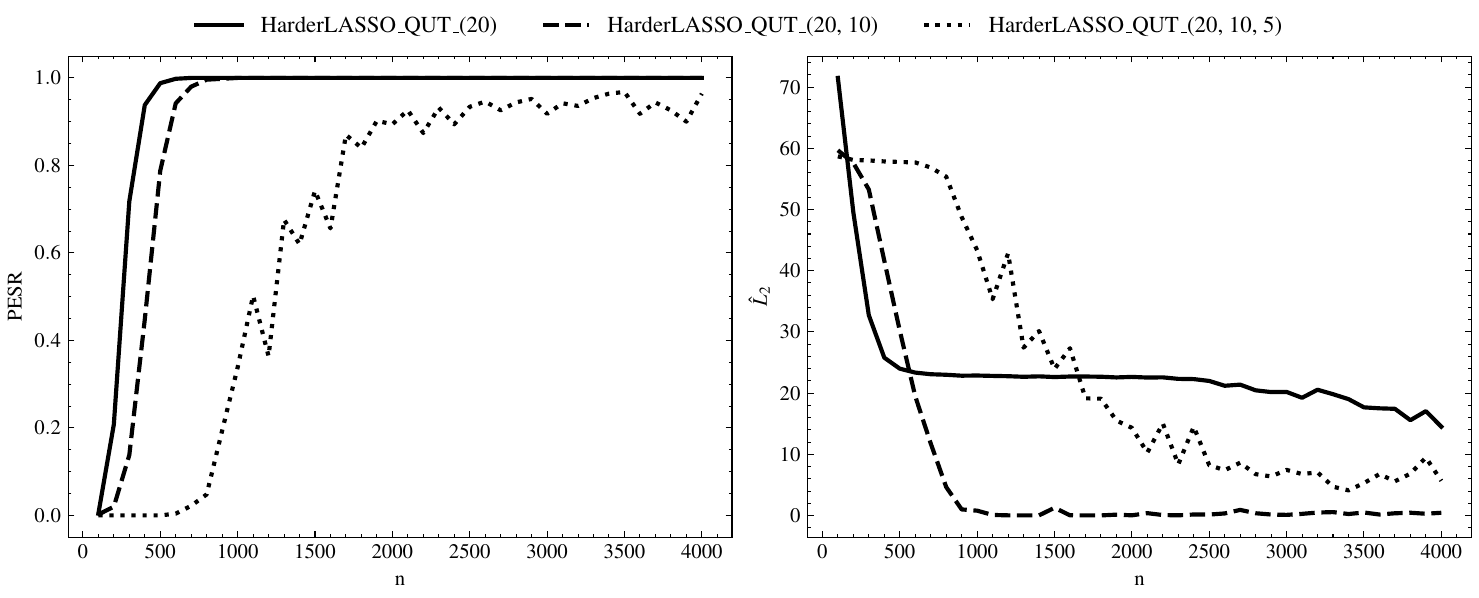}
    \caption{Results of Monte Carlo simulations ($m = 500$ runs) for the highly nonlinear setting. Generalization is evaluated on an independent test set of size $n = 1000$.}
    \label{fig:hardernonlinear_simu_results}
\end{figure}

Figure \ref{fig:hardernonlinear_simu_results} displays the Monte Carlo results. We observe that the 1 hidden layer ANN has a rapidly growing PESR (left), but its corresponding predictive performance (right) plateaus at a higher level since it does not have enough flexibility to match the full nonlinearity of the association. The 2 hidden layer ANN also has a rapid PESR increase and its predictive performance does not plateau but keeps on improving eventually reaching the minimum. The 3 hidden layer ANN has a slower PESR increase and its predictive performance nearly matches that of the 2 hidden layer ANN when n gets large.

\section{Real-world performance for the classification task}\label{sct:real_world}
We apply the five non linear learners described earlier (adapted for classification) on several “real-world” datasets detailed in Table \ref{tab: classification datasets}, which lists the number of instances $n$, number of features $p$ and the number of classes $m$. To quantify the accuracy relative to the number of needles identified by the learners, we average the models performance over $50$ resamplings to ensure stability and reliability in our findings. For each resampling, the dataset is randomly partitioned, with two-thirds of the data allocated for training and one-third reserved for testing. Mean imputation is employed for missing values. If separate training, testing, and validation sets are provided, these sets are concatenated to form the dataset before partitioning. 

\begin{table}[!ht]
\centering
\begin{tabular}{llrrr}
\toprule
\textbf{Type} & \textbf{Dataset} & \textbf{n} & \textbf{p} & \textbf{m}\\
\midrule
\multirow{3}{*}{Bioinformatics}
& Aids        & 2139 &  23   & 2 \\
& Breast Cancer    &  569 & 30  & 2 \\
& DNA    & 3186 &180 &3 \\
\midrule
\multirow{3}{*}{Image}
& USPS        & 9298 &  256   & 10 \\
& COIL20    &  1440 & 1024  & 20 \\
& Statlog & 6435 & 36 & 7 \\
\midrule
\multirow{2}{*}{Text}
& BASEHOCK  &  1993 &   4862  & 2 \\
& PCMAC       & 1943 &   3289  & 2 \\
\midrule
\multirow{2}{*}{Business}
& Bankruptcy    & 6819 & 95  & 2 \\
& Spambase     & 4601 & 57 & 5 \\
\midrule
\multirow{1}{*}{Speech}
& Isolet    & 1560 & 617  & 26 \\
\midrule
\multirow{3}{*}{Other}
& Wine    & 178 & 13 & 3 \\
& Dry Bean & 13611 & 16 & 7 \\
& HAR & 10299 & 561 & 6 \\
\bottomrule
\end{tabular}
\caption{Classification datasets with some key characteristics.}
\label{tab: classification datasets}
\end{table}

Table \ref{tab: results classification} summarizes the classification results. It is very clear that the QUT-based methods achieve the most aggressive feature selection across all datasets, reducing dimensionality to fewer than 20 features even on very large-scale problems, while still maintaining competitive levels of accuracy. In particular, for comparable levels of accuracy, \texttt{HarderLASSO\_QUT} is consistently more frugal, selecting fewer features than any alternative. Between the two QUT penalties, \texttt{HarderLASSO\_QUT} is slightly more sparse than \texttt{LASSO\_QUT}, but keeping very similar predictive performances.

When compared to competing approaches, our models stand out for their feature selection efficiency. Both the cross-validated \texttt{LASSO\_CV} and the neural approach \texttt{LassoNet}, while accurate, tend to retain a large fraction of the inputs; an inherent limitation of cross-validation which prioritizes predictive performance over sparsity. The considered ensemble models achieve strong accuracy with moderate dimensionality reduction, but they still select more features than the QUT-based approaches. Their competitive performance can be attributed to their structural design, which averages across a large collection of base learners. We anticipate that our QUT-based models could further close the predictive gap if extended with mechanisms such as majority voting or deeper architectures, while preserving their strong advantages in feature parsimony.

\begin{table}[!ht]
\centering
\resizebox{\columnwidth}{!}{
\begin{tabular}{l
  r
  rrr
  rr}
\toprule
& {\textbf{Linear}} & \multicolumn{3}{c}{\textbf{ANN}} & \multicolumn{2}{c}{\textbf{Ensemble}} \\
\cmidrule(lr){2-2}\cmidrule(lr){3-5}\cmidrule(lr){6-7}
Dataset
& LASSO\_CV & LassoNet & LASSO\_QUT & Harder\_QUT & RandomForest & XGBoost \\
\midrule
Aids & 21.2 (86.4) & 22.1 (87.7) & 2.8 (86.4) & 3.2 (88.1) & 4.2 (84.8) & 6.6 (88.8) \\
Breast Cancer & 15.9 (97.0)& 25.1 (96.9) & 4.2 (95.5) & 2.1 (94.3) & 21.5 (95.7) & 9.2 (95.7)\\
DNA & 106.1 (95.5) & 60.9 (95.1) & 11.4 (93.7) & 8.5 (91.5) & 84.5 (95.5) & 15.7 (94.2) \\
\midrule
USPS & 251.5 (94.0) & 256.0 (95.1) & 53.9 (90.6) & 17.5 (88.1) & 212.4 (96.3)& 109.0 (95.6) \\
COIL20 & 1021.7 (98.1) & 1012.0 (97.9) & 29.2 (91.2) & 9.4 (89.2) & 946.9 (99.8) & 114.7 (98.3) \\
Statlog & 36.0 (85.8) & 36.0 (88.0) & 22.4 (85.7) & 4.1 (85.0) & 36.0 (91.2) & 35.7 (91.4) \\
\midrule
BASEHOCK & 3839.9 (97.7) & 4247.6 (95.9) & 26.7 (83.2) & 13.7 (84.7) & 50.84 (94.6) & 54.4 (94.8) \\
PCMAC & 1907.0 (90.0) & 264.2 (89.9) & 18.1 (80.3) & 11.8 (81.7) & 14.4 (84.8) & 31.0 (89.2) \\
\midrule
Bankruptcy & 20.5 (96.8) & 50.3 (96.8) & 6.7 (96.7) & 2.7 (96.8) & 21.7 (97.0) & 10.1 (96.7) \\
Spambase & 55.4 (92.3) & 56.6 (93.5) & 24.9 (90.5) & 16.5 (91.4) & 19.1 (94.5) & 22.9 (94.8) \\
\midrule
Isolet & 615.0 (95.9) & 617.0 (94.4) & 35.4 (83.1) & 15.2 (81.3) & 603.2 (94.4) & 98.1 (92.7) \\
\midrule
Wine & 10.2 (97.2) & 11.5 (97.1) & 4.9 (96.0) & 2.2 (89.6) & 13.0 (97.7) & 7.0 (96.1) \\
Dry Bean & 16.0 (92.4) & 16.0 (92.3) & 8.7 (92.3) & 5.9 (92.3) & 16.0 (92.2) & 13.0 (92.6) \\
HAR & 555.1 (98.2) & 560.6 (97.7) & 32.2 (96.1) & 8.6 (93.0) & 468.3 (97.8) & 61.9 (98.4) \\
\bottomrule
\end{tabular}
}
\caption{Number of selected features with (\% accuracy) on the classification datasets. Metrics averaged over $50$ resamplings.}
\label{tab: results classification}
\end{table}

\section{Conclusion}\label{sct:conclusions}
The proposed learner HarderLASSO, like in compressed sensing, seems to achieve a phase transition in the probability of exact support recovery. The key ingredient to this success is the derivation of the quantile universal threshold under pure noise, along with an efficient optimization scheme. The new paradigm can be ported to other learners, other tasks, other probability laws to make them more frugal and interpretable. Gaining more insight on the location and sharpness of the phase transition calls for future theoretical developments.

HarderLASSO, despite not being an ensemble method and relying on a single regularization parameter that requires no validation set for its tuning, performs remarkably well compared to state-of-the-art learners. It achieves competitive accuracy while selecting a much smaller subset of truly relevant features. Beyond its performance, one of HarderLASSO most notable strengths is its user-friendliness: the workflow is essentially ‘click start and wait,’ as the model autonomously determines all necessary parameters. This stands in contrast to many existing feature-selection approaches, which typically require the user to specify in advance the desired level of sparsity or to impose an arbitrary threshold. In our case, the model identifies the relevant feature set directly, without such manual intervention, making it both powerful and accessible.

\appendix

\section{Proof of Theorem~\ref{th:rho1D}}

Call $f(\theta;y)$ the cost function \eqref{eq:1D} for given $\lambda>0$ and $\nu\in(0,1]$. We consider the case $y>0$. To create two local minima with the same cost value (one at zero and the other at some positive value $\theta$) for some high enough value $y$, we want to solve over  $y>0$ and $\theta>0$
\begin{eqnarray*}
	\left \{
	\begin{array}{l}
		f(0; y)=f(\theta; y)\\
		f'(\theta;y)=0
	\end{array}
	\right.
	&\Leftrightarrow&
	\left \{
	\begin{array}{l}
		y^2= (\theta-y)^2+ 2 \lambda  \frac{\theta}{1+\theta^{1- \nu}}\\
		\theta-y+ \lambda  \frac{1+\nu \theta^{1- \nu}}{(1+\theta^{1- \nu})^2}=0
	\end{array}
	\right . \\
	&\Leftrightarrow&
	\left \{
	\begin{array}{l}
		y= \theta/2+  \lambda  \frac{1}{1+\theta^{1- \nu}}\\
		y=\theta+ \lambda  \frac{1+\nu \theta^{1- \nu}}{(1+\theta^{1- \nu})^2}
	\end{array}
	\right . \\
	&\Leftrightarrow&
	\left \{
	\begin{array}{l}
		\theta^{2-\nu}+2\theta + \theta^\nu + 2 \lambda(\nu-1)=0 \\
		y=\theta/2+  \lambda  \frac{1}{1+\theta^{1- \nu}}
	\end{array}
	\right . ,
\end{eqnarray*}
where the threshold is $ \varphi(\lambda,\nu):=y$ and the jump is $ \kappa(\lambda,\nu):=\theta$.
To numerically   solve the first equation  of the system  over $\theta>0$ for $\nu \in (0,1]$ and $\lambda>0$ fixed, note that
\begin{eqnarray*}
	&\theta^{2-\nu}+2\theta + \theta^\nu + 2 \lambda(\nu-1)=0\\
	&\Leftrightarrow ~~~~~\left( \theta^{1-\frac{\nu}{2}}+\theta^{\frac{\nu}{2}}\right)^2=2 \lambda(1-\nu)\\
	&\Leftrightarrow ~~~~~  \theta^{1-\frac{\nu}{2}}+\theta^{\frac{\nu}{2}}=\sqrt{2 \lambda(1-\nu)}.
\end{eqnarray*}
And since $\theta^{1-\frac{\nu}{2}}+\theta^{\frac{\nu}{2}} \geq 2 \sqrt{\theta^{1-\frac{\nu}{2}} \cdot \theta^{\frac{\nu}{2}}}=2\sqrt{\theta}$  (inequality of arithmetic and geometric means),  we must have that $\theta \in \left(0, \frac{\lambda(1-\nu)}{2} \right]$.

\section{The zero-thresholding function}\label{app:zero-thresh}

This appendix provides the derivation of the zero-thresholding function $\lambda_0({\cal X}, {\cal Y})$ of a learner defined as a solution to~\eqref{eq:CS2}.

\subsection{Preliminaries}
We begin by analyzing the behavior of a feature selection compatible MLP when $\boldsymbol{\theta}^{(1)} = W_1 = \mathbf{0}$.

\begin{property}[Null model behavior]\label{prop:nullmodel}
Let $\mu_{(\boldsymbol{\theta}^{(1)},\boldsymbol{\theta}^{(2)})}$ be an $L$-layer MLP. If $\boldsymbol{\theta}^{(1)} = W_1 = \mathbf{0}$, then for every choice of $\boldsymbol{\theta}^{(2)}$ the network output is the constant function:
\begin{equation}
\mu_{(\mathbf{0},\boldsymbol{\theta}^{(2)})}({\mathbf{x}}) =\mathbf{c}_{\boldsymbol{\theta}^{(2)}},
\end{equation}
where $\mathbf{c}_{\boldsymbol{\theta}^{(2)}} = S_L \circ S_{L-1} \circ \cdots\circ S_2(\sigma(\mathbf{b}_1))$ is independent of $\bf x$.
\end{property}
Under this null regime ($\boldsymbol{\theta}^{(1)}=\bf 0$), choosing the maximum likelihood estimate (MLE) of the constant predictor is the sample mean for regression and the class‐probability average for cross‐entropy classification. In order to characterize the entire family of deeper-layer parameters that achieve this optimal constant, we make the following definition:
\begin{definition}[Null‐model parameter set]\label{def:null_param_set}
Let $\hat{\mathbf{c}}^{\rm MLE}=\arg\min_{c\in\R^m}\mathcal L_n\bigl(c\mathbf1_n;{\cal Y},{\cal X}\bigr)$. Define
    $\Theta_0^{(2)}=\left\{\boldsymbol{\theta}^{(2)}: \mathbf{c}_{\boldsymbol{\theta}^{(2)}} = \hat{\mathbf{c}}^{\rm MLE} \right\}$.
\end{definition}
So a trained network solution to~\eqref{eq:CS2} with $ \hat{\boldsymbol{\theta}}^{(1)}=\mathbf{0}$ outputs $\mu_{(\mathbf0,\hat{\boldsymbol{\theta}}^{(2)})}({\mathbf{x}})
= \hat{\mathbf{c}}^{\rm MLE}$ for every $\hat{\boldsymbol{\theta}}^{(2)}\in\Theta_0^{(2)}$.


\subsection{Existence}

The quantile universal threshold of Definition~\ref{def:QUT} requires the existence of a so-called zero-thresholding function.

\begin{definition}[Zero-thresholding function]\label{def:zero-thresh}
Let $P$ be a sparsity-inducing penalty of the optimization problem~\eqref{eq:CS2} as defined in Definition \ref{def:sparsindupen}. The infimum of the $\lambda$'s for which any $(\hat{\boldsymbol{\theta}}^{(1)}, \hat{\boldsymbol{\theta}}^{(2)}) = (\mathbf{0}, \hat{\boldsymbol{\theta}}^{(2)})$ are local minimizers defines the zero-thresholding function $\lambda_0({\cal X}, {\cal Y})$. 
\end{definition}

The following theorem proves the existence of $\lambda_0$ under a condition on the sparsity inducing penalty.

\begin{theorem}[Zero-thresholding function characterization]\label{thm:infnormgrad}
  Let $P$ be a sparsity-inducing penalty satisfying $\lim_{\epsilon \to 0^+} \frac{P(\epsilon \boldsymbol{\theta}, \lambda)}{\epsilon} = \lambda \|\boldsymbol{\theta}\|_1$ for any non-zero matrix $\boldsymbol{\theta}$. Define $g({\cal Y}, {\cal X}, \boldsymbol{\theta}^{(2)}) = \nabla_{\boldsymbol{\theta}^{(1)}}\mathcal{L}_n(\mu_{(\boldsymbol{\theta}^{(1)},
  \boldsymbol{\theta}^{(2)})}; {\cal Y}, {\cal X})\big|_{\boldsymbol{\theta}^{(1)}=\mathbf{0}}$. Then
  \begin{equation}
  \lambda_0({\cal X}, {\cal Y}) =
  \sup_{\boldsymbol{\theta}^{(2)} \in \Theta_0^{(2)}} \left\| g({\cal Y},{\cal X},\boldsymbol{\theta}^{(2)})\right\|_{\mathrm{max}},
  \end{equation}
  where $\|A\|_{\mathrm{max}} = \max_{i,j}|a_{ij}|$ is the entrywise
  maximum absolute value.
\end{theorem}
The condition on the sparsity inducing penalty simply requires that $P$ exhibits the familiar "V-shaped" cusp of $|\theta|$ at the origin; its derivative jumps abruptly from $-\lambda$ to $+\lambda$, which provides the sharp edge needed to drive small coefficients exactly to zero. The $\ell_1$, SCAD, MCP penalties and along with the proposed penalty  $P_\nu$ for $0<\nu<1$ defined in~\ref{eq:Pnu} all share this cusp property. On the contrary,  the non-convex $\ell_\nu$ Subbotin penalty \citep{SardySLIC09, Woodworth2016} is sparsity-inducing but has an infinitely steep derivative at the origin when $\nu \in (0,1)$, so that is does not satisfy the condition of Theorem~\ref{thm:infnormgrad}. 

\textbf{Proof \ref{thm:infnormgrad}.} Let $\boldsymbol{\theta}^0 = (\mathbf{0}, {\boldsymbol{\theta}}^{(2)})$ with ${\boldsymbol{\theta}}^{(2)}\in\Theta_0^{(2)}$ of Definition~\eqref{def:null_param_set}. The optimality condition implies that the gradient of the loss with respect to ${\boldsymbol{\theta}}^{(2)}$ is zero. Let $W_1^{(\text{dir})}$ be a direction matrix for the weights with $\|W_1^{(\text{dir})}\|_1=1$. We define a perturbed parameter vector $\boldsymbol{\theta}^\epsilon = (\epsilon W_1^{(\text{dir})}, {\boldsymbol{\theta}}^{(2)})$ where $\epsilon$ is a small scalar. Since the loss function is twice differentiable with respect to $W_1$ around $\boldsymbol{\theta}^0=(\mathbf{0}, {\boldsymbol{\theta}}^{(2)})$, applying Taylor’s theorem we have
\begin{equation*}
\begin{aligned}
& \left|\mathcal{L}_n(\mu_{\boldsymbol{\theta}^\epsilon}; {\cal Y}, {\cal X})-\mathcal{L}_n(\mu_{\boldsymbol{\theta}^0}; {\cal Y}, {\cal X})\right| \\
& =\left|\left\langle\nabla_{W_1}\mathcal{L}_n(\mu_{\boldsymbol{\theta}^0}; {\cal Y}, {\cal X}), \epsilon W_1^{(\text{dir})}\right\rangle+o\left(\epsilon^2\left\|W_1^{(\text{dir})}\right\|_1\right)\right| \\
& \leq|\epsilon|\left|\left\langle\nabla_{W_1}\mathcal{L}_n(\mu_{\boldsymbol{\theta}^0}; {\cal Y}, {\cal X}), W_1^{(\text{dir})}\right\rangle\right|+o\left(\epsilon^2\right)\\
& \leq |\epsilon|\left\|\nabla_{W_1} \mathcal{L}_n(\mu_{\boldsymbol{\theta}^0}; {\cal Y}, {\cal X})\right\|_{\infty}+o\left(\epsilon^2\right).
\end{aligned}
\end{equation*}
Therefore we get 
$\mathcal{L}_n(\mu_{\boldsymbol{\theta}^\epsilon}; {\cal Y}, {\cal X}) \geq \mathcal{L}_n(\mu_{\boldsymbol{\theta}^0}; {\cal Y}, {\cal X})-|\epsilon|\left\|\nabla_{W_1} \mathcal{L}_n(\mu_{\boldsymbol{\theta}^0}; {\cal Y}, {\cal X})\right\|_{\infty}+o\left(\epsilon^2\right)$.
It follows that the regularized loss satisfies
$$
\begin{aligned}
& \mathcal{L}_n(\mu_{\boldsymbol{\theta}^\epsilon}; {\cal Y}, {\cal X})+ P_\nu\left(\epsilon W_1^{(\text{dir})}, \lambda\right) \\
& \geq \mathcal{L}_n(\mu_{\boldsymbol{\theta}^0}; {\cal Y}, {\cal X})-|\epsilon|\left\|\nabla_{W_1} \mathcal{L}_n(\mu_{\boldsymbol{\theta}^0}; {\cal Y}, {\cal X})\right\|_{\infty}+ P_\nu\left(\epsilon W_1^{(\text{dir})}, \lambda\right)+o\left(\epsilon^2\right).
\end{aligned}
$$
If $P(\epsilon W_1^{(\text{dir})}, \lambda) = \epsilon\lambda \|W_1^{(\text{dir})}\|_1 + o(\epsilon)$, the change in regularized loss is
\begin{equation*}
    \mathcal{L}_n(\mu_{\boldsymbol{\theta}^\epsilon}; {\cal Y}, {\cal X})+ P_\nu\left(\epsilon W_1^{(\text{dir})}, \lambda\right)\geq \mathcal{L}_n(\mu_{\boldsymbol{\theta}^0}; {\cal Y}, {\cal X})+|\epsilon| \left( \lambda - \left\|\nabla_{W_1} \mathcal{L}_n(\mu_{\boldsymbol{\theta}^0}; {\cal Y}, {\cal X})\right\|_{\infty} \right) + o(\epsilon).
\end{equation*}
If we assume that $(\lambda - \left\|\nabla_{W_1} \mathcal{L}_n(\mu_{\boldsymbol{\theta}^0}; {\cal Y}, {\cal X})\right\|_{\infty}) > C$ for some constant $C > 0$, 
\begin{equation*}
    \mathcal{L}_n(\mu_{\boldsymbol{\theta}^\epsilon}; {\cal Y}, {\cal X})+ P_\nu\left(\epsilon W_1^{(\text{dir})}, \lambda\right)\geq \mathcal{L}_n(\mu_{\boldsymbol{\theta}^0}; {\cal Y}, {\cal X})+|\epsilon|C+o(\epsilon)\geq \mathcal{L}_n(\mu_{\boldsymbol{\theta}^0}; {\cal Y}, {\cal X})
\end{equation*}
for $|\epsilon|$ small enough. Thus, the cost function with our choice of $\lambda$ indeed has a local minimum at $\boldsymbol{\theta}^0$.

\subsection{Explicit Computation for MLPs (Lemma \ref{lem:lambda_regclass})}
We now compute the explicit form of $\lambda_0$ for feature selection compatible MLPs.

\textbf{For regression} with the square-root $\ell_2$ loss
\begin{equation*}
    \mathcal{L}_n(\mu_{\boldsymbol{\theta}}; {\cal Y}, {\cal X})=\left\|\cal{Y}-\mu_{\boldsymbol{\theta}}({\cal X})\right\|_2=\sqrt{\sum_{k=1}^n\left(y_k-\left(c+\mathbf{w}_L \sigma\left(\mathbf{b}_{L-1}+W_{L-1} \sigma\left(\cdots \sigma\left(\mathbf{b}_1+W_1 \mathbf{x}_k\right)\right)\right)\right)^2\right.},
\end{equation*}
where $y_k \in \mathbb{R},\ \mathbf{x}_k\in\mathbb R^{p_1\times 1},\ c \in \mathbb{R},\ \mathbf{w}_L \in \mathbb{R}^{1\times p_L},\ W_l \in \mathbb{R}^{p_{l+1} \times p_l}$ and $\mathbf{b}_l \in \mathbb{R}^{p_{l+1} \times 1}$ for $l=1, \ldots, L-1$. At $\boldsymbol{\theta}^{(1)}=W_1=\mathbf{0}$, the least squares problem is solved for $\hat{\boldsymbol{\theta}}^{(2)}\in\Theta_0^{(2)}$ so that $\mu_{(\mathbf 0, \hat{\boldsymbol{\theta}}^{(2)})}({\bf x}_k)=\bar{\cal Y}$, the average of the training set responses, for $k=1,\dots, n$. Defining, $\mathbf{z}_{l}^{(k)}=\mathbf{b}_{l}+W_{l}\mathbf{a}_{l}^{(k)}$, $\mathbf{a}_{l}^{(k)}=\sigma(\mathbf{z}_{l-1}^{(k)})$, $\mathbf{a}_{1}^{(k)}=\mathbf{x}_k$ and $g(\mathcal{Y}, \mathcal{X}, \boldsymbol{\theta}^{(2)})=\nabla_{\boldsymbol{\theta}^{(1)}}\mathcal{L}_n(\mu_{({\boldsymbol \theta}^{(1)}, {\boldsymbol \theta}^{(2)})}; {\cal Y}, \mathcal{X}) \Big\vert_{\boldsymbol{\theta}^{(1)}=\mathbf{0}}$, some elementary calculation yields
\begin{equation*}
    g(\mathcal{Y}, \mathcal{X}, \hat{\boldsymbol{\theta}}^{(2)}) = \frac{1}{\|{\mathcal{Y}-\bar{\mathcal{Y}}\mathbf{1}_n}\|_2}\left[-D_1W_2^{\rm T}D_2W_3^{\rm T}\ldots D_{L-1}\mathbf{w}_L^{\rm T}\right]\sum_{k=1}^{n}(y_k-\bar{\cal{Y}})\mathbf{x}_k^{\rm T},
\end{equation*}
where $D_l=\operatorname{diag}(\sigma'(\mathbf{z}_l))$ and, when $\boldsymbol{\theta}^{(1)}=\mathbf{0}$, each $\mathbf{z}_{l}^{(k)}=\mathbf{z}_{l}$ (independent of $k$). Choosing $c=\mathcal{\bar{Y}}-\mathbf{w}_L\sigma\left(\mathbf{b}_{L-1}+W_{L-1}\sigma\left(\dots\sigma\left(\mathbf{b}_1\right)\right)\right)$ ensures $\boldsymbol{\theta}^{(2)}\in\Theta_0^{(2)}$ no matter the biases $\mathbf{b_1}, \dots, \mathbf{b}_{L-1}$ and the weights $W_2, \dots \mathbf{w}_L$. Let $\kappa:=\sup_{t}|\sigma'(t)|$, we can choose biases $\mathbf{b_1}, \dots, \mathbf{b}_{L-1}$ so that $D_l=\kappa\operatorname{Id},\ l=1,\dots, L-1$. Hence, $\sup_{\mathbf{b}_i}\|D_1W_2^{\rm T}\dots D_{L-1}\mathbf{w}_L^{\rm T}\|_{\rm max}=\kappa^{L-1}\|W_2^{\rm T}W_3^{\rm T}\dots\mathbf{w}_L^{\rm T}\|_{\rm max}$ and,
\begin{gather*}
    \sup_{\boldsymbol{\theta}^{(2)}\in\Theta_0^{(2)}}\left\|g(\mathcal{Y}, \mathcal{X}, \boldsymbol{\theta}^{(2)})\right\|_{\rm max} = \frac{\kappa^{L-1}}{\|{\mathcal{Y}-\bar{\mathcal{Y}}\mathbf{1}_n}\|_2}\sup_{(W_2, \dots, \mathbf{w}_L)\in \mathcal{D}_L}\left\|\left(W_2^{\rm T}W_3^{\rm T}\ldots \mathbf{w}_L^{\rm T}\right)\sum_{k=1}^{n}(y_k-\bar{\mathcal{Y}})\mathbf{x}_k^{\rm T}\right\|_{\rm max}\\
    =\frac{\kappa^{L-1}\|X^{\rm T}({\mathcal{Y}-\bar{\mathcal{Y}}\mathbf{1}_n})\|_\infty}{\|{\mathcal{Y}-\bar{\mathcal{Y}}\mathbf{1}_n}\|_2}\sup_{(W_2, \dots, \mathbf{w}_L)\in \mathcal{D}_L}\max _i\left\{\left|\mathbf{w}_L W_{L-1} \ldots W_{2 ; :i}\right|\right\},
\end{gather*}
where 
\begin{equation*}
    \mathcal{D}_L=\left\{\left(W_2, \ldots, W_{L-1}, \mathbf{w}_L\right):\left\|W_{h ; i:}\right\|_2 =1\text{ for }i \in\left\{1, \ldots, p_{h+1}\right\}, h \in\{2, \ldots, L-1\},\left\|\mathbf{w}_L\right\|_2=1\right\}
\end{equation*}
is the set of weights matrices having $\ell_2$ normalized rows. Now, we concentrate on $\operatorname{supmax}_L = \sup _{(W_2, \dots, \mathbf{w}_L)\in \mathcal{D}_L} \max _i\left\{\left|\mathbf{w}_L W_{L-1} \ldots W_{2 ;: i}\right|\right\}$. When $L=2$, $\operatorname{supmax}_L=\sup _{\mathbf{w}_2\in \mathcal{D}_2} \max _i\left|w_{2, i}\right|=1$. When $L=3$, $\operatorname{supmax}_L=\sup _{(W_2, \mathbf{w}_3)\in \mathcal{D}_3}\max _i\left|\mathbf{w}_3 W_{2 ; :i}\right|$. Because $W_{2 ; :i}$ is the i-th column of $W_2$ and the $l_2$ norm of every row is $1$, the maximum is reached by setting one column of $W_2$ to $\mathbf{1}_{p_3}$ and setting its all other elements to zero. Then $\operatorname{supmax}_L=\max _{\|\mathbf{w}_3\|_2=1} \sum_{i=1}^{p_3} w_{3, i}=\sqrt{p_3}$ (Cauchy Schwarz inequality). Likewise, the result generalizes to $L$-layers with
\begin{equation*}
    \sup_{\boldsymbol{\theta}^{(2)}\in\Theta_0^{(2)}}\left\|g(\mathcal{Y}, \mathcal{X}, \boldsymbol{\theta}^{(2)})\right\|_{\rm max} = \kappa^{L-1}\sqrt{p_3p_4\dots p_{L}}\frac{\|\mathcal{X}^{\rm T}({\mathcal{Y}-\bar{\mathcal{Y}}\mathbf{1}_n})\|_\infty}{\|{\mathcal{Y}-\bar{\mathcal{Y}}\mathbf{1}_n}\|_2}.
\end{equation*}

\noindent\textbf{For classification} with the cross-entropy loss 
\begin{equation*}
    \mathcal{L}_n(\mu_{\boldsymbol{\theta}}; {\cal Y}, {\cal X})= - \sum_{k=1}^n \mathbf{y}_k^{\rm T}\log\left(\frac{\exp{\left(\mathbf{c}+\mathbf{W}_L \sigma\left(\cdots \sigma\left(\mathbf{b}_1+W_1 \mathbf{x}_k\right)\right)\right)}}{\sum_{t=1}^m\exp{\left(c_t+\mathbf{W}_{L;t:} \sigma\left(\cdots \sigma\left(\mathbf{b}_1+W_1 \mathbf{x}_k\right)\right)\right)}}\right),
\end{equation*}
where $\boldsymbol{y}_k\in\{0, 1\}^{m}$ is a one-hot vector $\mathbb R^{m\times 1}$,\ $\mathbf{x}\in\mathbb{R}^{p_1\times 1}$ $W_L\in\mathbb R^{m\times p_L}$, $W_l\in\mathbb R^{p_{l+1}\times p_l}$ for $l=1, \dots, L-1$ and $m$ is the number of classes. Using similar arguments as those for regression, one has the entrywise max-abs norm
\begin{equation*}
\sup_{\boldsymbol{\theta}^{(2)}\in\Theta_0^{(2)}}\left\|g(\mathcal{Y}, \mathcal{X}, \boldsymbol{\theta}^{(2)})\right\|_{\rm max} = \kappa^{L-1}\sup_{(W_2, \dots, W_L)\in \mathcal{D}_L}\left\|\left(W_2^{\rm T}W_3^{\rm T}\ldots W_L^{\rm T}\right)\sum_{k=1}^{n}(\bar{\mathcal{Y}}-\mathbf{y}_k)\mathbf{x}_k^{\rm T}\right\|_{\rm max}
\end{equation*}
where $\mathcal{D}_L$ is the set $\left\{\left(W_2, \ldots, W_L\right):\left\|W_{h ; i:}\right\|_2=1 \text{ for }i \in\{1, \ldots, p_{h+1}\},\ h \in\{2, \ldots, L\}\right\}$, where $W_{h ; i:}$ is the i-th row of $W_h$. Similarly, via choosing $W_{2 ;: i}=\mathbf{1}_{p_3 \times 1}^{\mathrm{T}}$, $W_{h; i:}^{\mathrm{T}}=1 / \sqrt{p_h}\mathbf{1}$ for $h=3, \ldots, L-1 ; i=1,2, \ldots, p_{h+1}$, and $W_{L; i:}^{\mathrm{T}}=1 / \sqrt{p_L} \operatorname{sgn}\left(\left(\sum_{k=1}^n x_{k, j}\left(\bar{\boldsymbol{\mathbf{y}}}-\boldsymbol{y}_k\right)\right)_i\right) \mathbf{1}$ for any $i \in\{1,2, \ldots, m\}$, we have
\begin{equation*}
\begin{split}
    \sup_{\boldsymbol{\theta}^{(2)}\in\Theta_0^{(2)}}\left\|g(\mathcal{Y}, \mathcal{X}, \boldsymbol{\theta}^{(2)})\right\|_{\rm max} &= \kappa^{L-1}\sqrt{p_3\dots p_L}\max_{1\leq j\leq p_1}\sum_{t=1}^m|\sum_{k=1}^n (\bar{\mathcal{Y}}-\mathbf{y}_k)\mathbf{x}_k^{\rm T}|\\
    &=\kappa^{L-1}\sqrt{p_3\dots p_L}\|\mathcal{X}^{\rm T}\mathcal{Y}\|_\infty.
\end{split}
\end{equation*}

\section{Simulations: FDR \& TPR}
\begin{figure}[h]
    \centering
    \includegraphics[width=\linewidth]{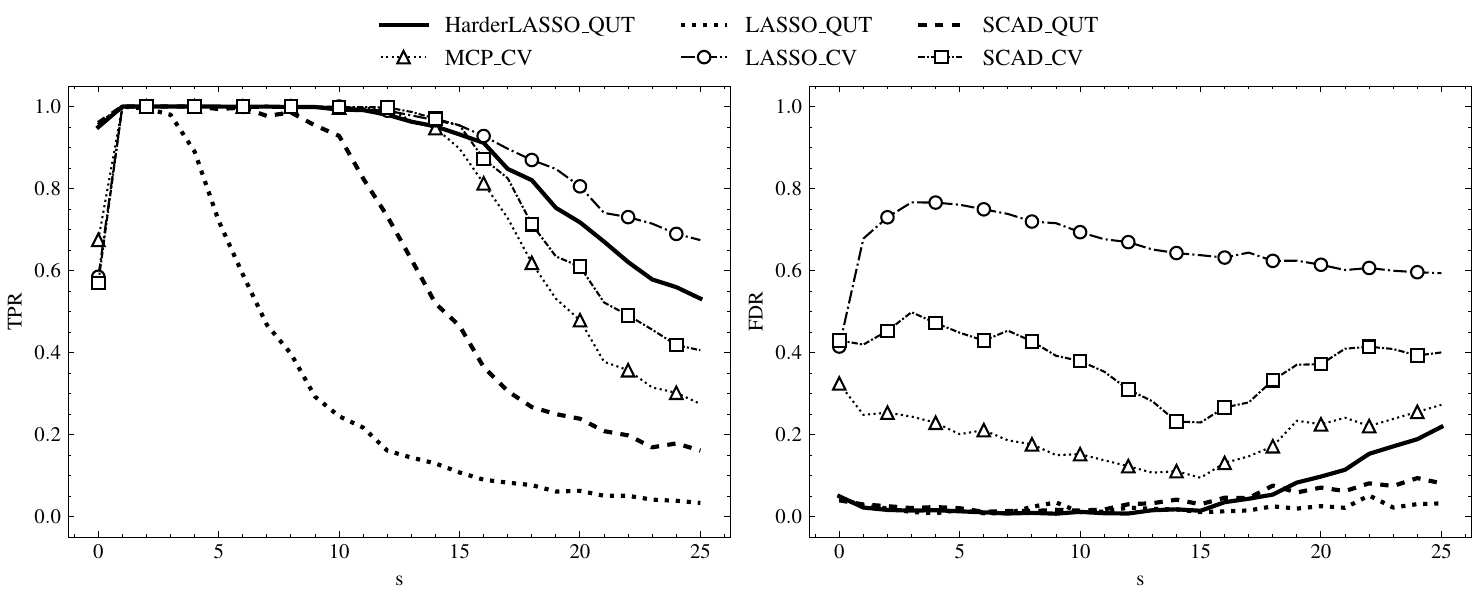}
    \caption{TPR and FDR in linear setting.}
\end{figure}

\begin{figure}[h]
    \centering
    \includegraphics[width=\linewidth]{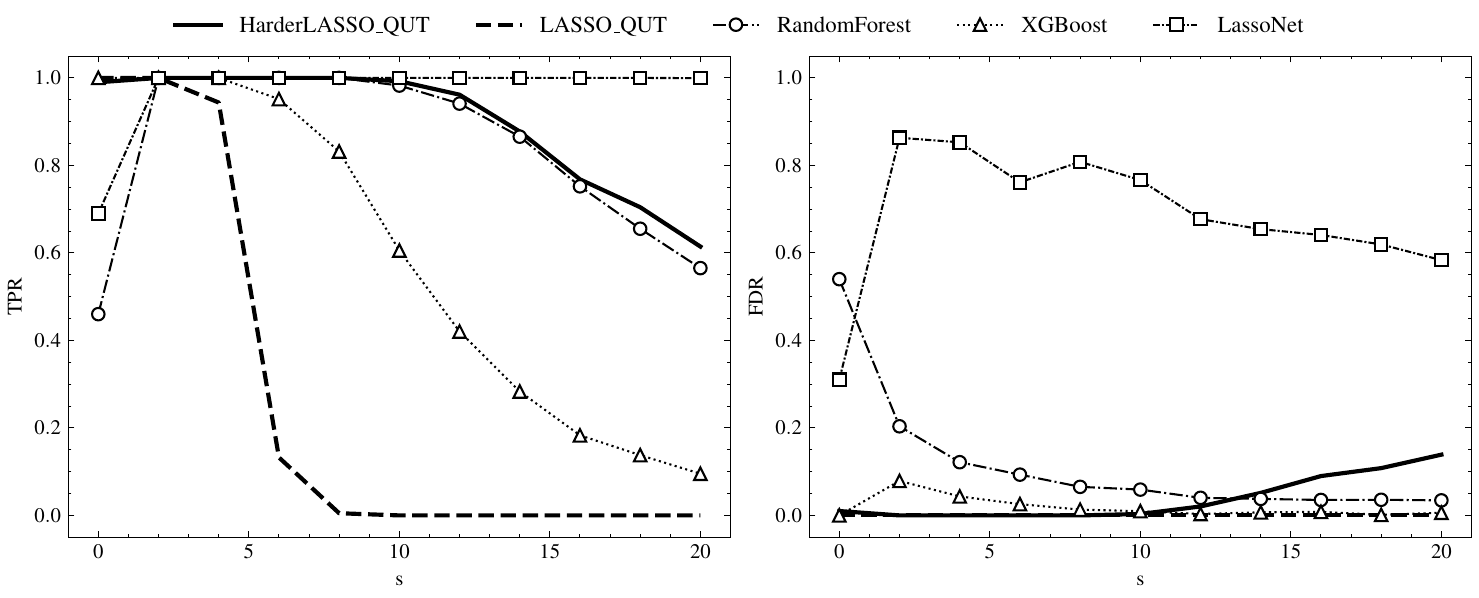}
    \caption{TPR and FDR in nonlinear setting.}
\end{figure}

\newpage

\bibliographystyle{plainnat}
\bibliography{article_bis}

\end{document}